\documentclass[
]{ceurart}

\usepackage{listings}
\usepackage{tabularx}
\usepackage{graphicx}

\begin{document}

\copyrightyear{2024}
\copyrightclause{Copyright for this paper by its authors.
  Use permitted under Creative Commons License Attribution 4.0
  International (CC BY 4.0).}

\conference{EKAW 2024: EKAW 2024 Workshops, Tutorials, Posters and Demos, 24th International Conference on Knowledge Engineering and Knowledge Management (EKAW 2024), November 26-28, 2024, Amsterdam, The Netherlands.}

\title{Named Entity Recognition in Historical Italian: The Case of Giacomo Leopardi's Zibaldone}

\author[1]{Cristian Santini}[%
orcid=0000-0001-7363-6737,
email=c.santini12@unimc.it,
]
\cormark[1]
\author[1]{Laura Melosi}[%
email=laura.melosi@unimc.it,
]
\cormark[1]
\author[2]{Emanuele Frontoni}[%
orcid=0000-0002-8893-9244,
email=emanuele.frontoni@unimc.it,
]
\cormark[1]
\address[1]{Department of Humanities, University of Macerata, Macerata, Italy}
\address[2]{Department of Political Sciences, Communication and International Relations, University of Macerata, Macerata, Italy}

\cortext[1]{Corresponding author.}

\begin{abstract}
  The increased digitization of world's textual heritage poses significant challenges for both computer science and literary studies. Overall, there is an urgent need of computational techniques able to adapt to the challenges of historical texts, such as orthographic and spelling variations, fragmentary structure and digitization errors. The rise of large language models (LLMs) has revolutionized natural language processing, suggesting promising applications for Named Entity Recognition (NER) on historical documents. In spite of this, no thorough evaluation has been proposed for Italian texts. This research tries to fill the gap by proposing a new challenging dataset for entity extraction based on a corpus of 19th century scholarly notes, i.e. Giacomo Leopardi's Zibaldone (1898), containing 2,899 references to people, locations and literary works. This dataset was used to carry out reproducible experiments with both domain-specific BERT-based models and state-of-the-art LLMs such as LLaMa3.1. Results show that instruction-tuned models encounter multiple difficulties handling historical humanistic texts, while fine-tuned NER models offer more robust performance even with challenging entity types such as bibliographic references.
\end{abstract}

\begin{keywords}
  Digital humanities \sep
  Historical documents \sep
  NER \sep
  Large language models \sep
  LLaMa \sep
  GliNER \sep
  Giacomo Leopardi \sep
  Zibaldone
\end{keywords}

\maketitle

\section{Introduction}

The increased digitization of the world's literary heritage opens new horizons for research in the field of Digital Humanities (DH).  This research area has to confront some  significant challenges that AI will face in the years to come. First of all, bibliographic resources are cultural products and, as such, they are subject to multiple interpretations and are described by libraries and research communities using various standards. This creates a wide variety of digitization approaches, which manifests in the multiplicity of digital edition projects and the creation of increasingly specific vocabularies and ontologies to describe these resources and their content~\cite{gaitanou_linked_2024}. From a digital perspective, this heterogeneity influences the ability of systems based on Machine Learning (ML) or Deep Learning (DL) techniques to understand and operate with this type of data. Nevertheless, the recent success of DL approaches in many fields opens up new possibilities for the DH. In specific, the advent of  Large Language Models (LLMs), such as GPT~\cite{brown_language_2020} and LLaMa~\cite{touvron2023llamaopenefficientfoundation} pushed researchers to analyze their performances on domain-specific texts such as historical multilingual documents \cite{gonzalez-gallardo_yes_2023, gonzalez-gallardo_leveraging_2024}. 

The integration of language models into the analysis of humanistic texts is emerging as a promising field of study for Information Extraction (IE) and Named Entity Recognition (NER). Aladağ~\cite{aladag_potential_2023} uses LLMs in her study on the computational analysis of Evliya Çelebi's travels to identify semantic and thematic patterns, overcoming the linguistic challenges of Ottoman Turkish. In the proposed approach, the authors use ChatGPT for NER of entities such as proper names, people, places, and other categories within the text to calculate the most frequent entities. Similarly, the project described by Spina~\cite{spina_biscari_2023} aims to demonstrate how the use of AI tools, such as Transkribus and ChatGPT, can revolutionize the digitization and accessibility of Italian historical archives. In their study, the authors use ChatGPT for zero-shot recognition of references to persons and locations inside the Princes of Biscari correspondence. Overall, these studies highlight both the benefits and challenges of applying LLMs to humanistic texts, opening new perspectives for DH regarding the specialization of these tools for historical and literary documents.

Building on the state of the art, this research aims to quantitatively analyze the results of various NER algorithms based on language models by creating a new dataset for NER on historical Italian from a publicly-available resource: DigitalZibaldone~\cite{stoyanova_working_2023}, a digital edition of Leopardi’s Zibaldone (1898). Written in several years by the Italian poet Giacomo Leopardi (1798 - 1837), the \textit{Zibaldone} is a collection of more than 4000 notes on several topics, including philosophy, science, linguistics, politics, history and philology. The digital edition of this work, made by scholars and technologists at Princeton University, is an interesting resource for the evaluation of entity extraction tools for two reasons: first, it contains thousands of manually annotated references to people, places and bibliographic works linked to VIAF\footnote{\url{https://viaf.org/}} and Wikidata\footnote{\url{https://www.wikidata.org/}} by domain-experts; secondly, the research notes of Giacomo Leopardi contained in this work represent a pre-digital example of an encyclopedic hyper-text, including ca. 10,000 internal and external references to historical figures, authors, literary works and other instances of humanistic knowledge~\cite{stoyanova_fragmentary_2013}. To summarise, the contributions of this study are the following:

\begin{itemize}
    \item to propose a new dataset for Named Entity Recognition and Linking in order to benchmark language models on historical humanistic texts;
    \item to understand how efficient LLMs can be in recognizing references to external entities in 19th century Italian scholarly notes and the actual challenges posed by these documents;
    \item to propose a new domain-specific NER model trained on historical humanistic texts;
    \item to provide a set of recommendations for integrating LLMs into the extraction of information from documents contained in archives of Italian authors.
\end{itemize}

This work is structured as follows. Section~\ref{sec:related_work} introduces the current state of the art with respect to LLMs and NER, with a focus on applications for historical texts. Section~\ref{sec:materials_and_methods} presents the dataset generated for the experiments, describes the different NER methodologies tested, and discusses in detail the evaluation metrics and tools used in the experiments. Section~\ref{sec:results} reports the values obtained in measuring the models' ability to recognize entities from the Zibaldone. Section~\ref{sec:discussion_and_conclusion} provides a detailed analysis of the results obtained and final recommendations on the use of the analyzed models in literary digitization projects. Additionally, this section outlines future extensions of this study.

\section{Related Work}
\label{sec:related_work}

The rise of LLMs has revolutionized NLP, enabling significant progress in tasks such as Named Entity Recognition (NER). Despite the impressive performance of LLMs for IE on modern texts \cite{wang_instructuie_2023, sainz_gollie_2024}, their application to historical texts remains underexplored. Historical documents present unique challenges, such as language evolution, orthographic variation, and digitization errors, which hinder the direct application of models designed for contemporary language \cite{ehrmann2021named}. Currently, a growing body of research has attempted to leverage LLMs for historical NER. González-Gallardo et al.~\cite{gonzalez-gallardo_yes_2023} explored the capabilities of ChatGPT for entity recognition in historical documents, applying it to diverse datasets such as NewsEye~\cite{hamdi_multilingual_2021}, hipe-2020~\cite{ehrmann_extended_2020} and AJMC~\cite{romanello_named_2024}. The study revealed that while LLMs could handle modern texts with some success, their performance on historical documents was inconsistent, particularly when dealing with the linguistic complexity of older texts. Problems such as entity overlap, code-switching between languages, and digitization errors, especially from optical character recognition (OCR), significantly degraded the model's accuracy. The authors concluded that task-specific training and better domain adaptation are required for LLMs to perform well on historical NER tasks.

Similarly, another study by González-Gallardo and colleagues~\cite{gonzalez-gallardo_leveraging_2024} evaluated open-source LLMs such as Llama \cite{touvron2023llamaopenefficientfoundation} and Mistral \cite{jiang_mistral_2023} on historical NER tasks across a series of historical datasets in several European languages including French, German and English. The findings underscored the difficulty of recognizing entities in noisy, digitized texts, with models frequently misclassifying or failing to recognize domain-specific entities such as literary works. Moreover, LLMs struggled with the structural inconsistencies common in historical documents, such as incomplete or fragmented texts. Despite advancements in prompt engineering and multi-turn interaction modes, the performance of these models lagged significantly behind fine-tuned NER models, emphasizing the need for more specialized tools tailored to historical document analysis.

The study of historical texts in non-Western languages has also yielded similar results. Tang et al.~\cite{tang_chisiec_2024} introduced the CHisIEC corpus for NER and Relation Extraction (RE) in ancient Chinese texts. The authors highlighted the linguistic diversity of ancient Chinese, which spanned multiple dynasties and included substantial variation in language use across time periods. Despite the use of specialized pre-trained language models fine-tuned on ancient Chinese, LLMs struggled with recognizing entities and faced particular challenges with the recognition of historical official positions and book titles. This study further emphasized the need for domain-specific language models, particularly in historical contexts where entity recognition is hampered by the lack of standardization in entity annotations and temporal variations in language.

In a different context, Spina~\cite{spina_biscari_2023} explored the use of AI tools like Transkribus and ChatGPT for transcribing and extracting entities from historical correspondence in the Biscari Archive, in Sicily. This project faced several difficulties due to the handwritten nature of the documents and their digitization into machine-readable formats. The integration of AI models like ChatGPT for zero-shot NER enabled the extraction of persons and locations from the transcribed texts, however the authors did not provide a thorough evaluation of the results and only a qualitative evaluation was provided. Lastly, Aladağ~\cite{aladag_potential_2023} applied the same chatbot to analyze Ottoman historical texts, particularly Evliya Çelebi’s travelogue. While ChatGPT exhibited potential in extracting thematic and semantic patterns, the model struggled with the linguistic complexities of Ottoman Turkish, which incorporates elements from Arabic and Persian. The study concluded that current LLMs are ill-equipped to handle the intricacies of historical languages, and the lack of tailored models for such contexts limits the effectiveness of NER and other NLP tasks.

In all these studies, a common set of challenges emerges: noisy, digitized text, linguistic variation across time, and entity ambiguity or overlap. These challenges are particularly acute in historical texts, which lack the consistency and structure found in contemporary language corpora. While LLMs have shown promise in recognizing named entities in historical documents, their performance lags behind task-specific models, especially in the presence of OCR errors and complex, multilingual contexts. These limitations highlight the importance of developing models that are not only trained on historical language corpora but also capable of handling the variability and fragmentation inherent in such documents.

Notably absent from the existing body of research is a comprehensive study of NER in historical Italian texts, due to the lack of open and reusable datasets extracted from non-contemporary documents. Italian historical documents, ranging from medieval manuscripts to Renaissance~\cite{santini2022knowledge} and post-Renaissance texts, exhibit significant orthographic and syntactic variations. Despite the need for accurate NER systems to support the DH in Italy, no large-scale, domain-specific corpus of texts currently exists for testing the performance of LLMs on historical Italian. Therefore, there is an urgent need to develop benchmarks and datasets for quantitative studies on the efficiency of LLMs in this domain. Such efforts would pave the way for more accurate entity recognition, fostering a deeper understanding of Italy's historical and cultural heritage.

\section{Materials and Methods}
\label{sec:materials_and_methods}

\subsection{Dataset}
The objective of this research is to compare the performance of different open-source LLMs for Named Entity Recognition (NER) on historical Italian literary texts. To carry out this research, a new dataset for NER and Entity Linking (EL), was created based on a publicly available online resource: DigitalZibaldone\footnote{\url{https://digitalzibaldone.net/}}. This resource is structured as a website where each note of Leopardi can be accessed on a specific URI and is encoded using HTML. For example, by searching the page identified with URI \url{https://digitalzibaldone.net/node/p2721_1} the user can visualize the first note in page 2721 of Leopardi's diary as an HTML page. In this edition, references to persons, locations and bibliographic works are encoded as links which redirect to the corresponding Wikidata entity, whenever present. If an entity is not present in Wikidata, an identifier will be provided using VIAF, the Virtual International Authority File.

Initially, 260 notes from the Zibaldone (pp. 2700-3000) written by Leopardi in 1823, one of the writer's most productive years, were identified as the evaluation dataset. Once these notes were identified, the HTML source code for each item was downloaded from the respective URI on the DigitalZibaldone platform using a web scraping algorithm developed in \texttt{Python}. A subsequent program, aimed at parsing the HTML file, was used to extract from each file the references to people, places and works defined through the hyperlinks present in the text. This algorithm produced two final \texttt{CSV} files: the first containing the text of each paragraph identified by an ID and cleaned of HTML formatting elements; the second containing all the annotations present in the 260 notes with the document ID, the position in the text of the annotated characters, the type of entity annotated (\texttt{PER}, \texttt{LOC} or \texttt{WORK}), and the link to the respective Wikidata or VIAF element. Examples of the two files contained in the evaluation dataset are shown in Table \ref{tab1} and Table \ref{tab2}.

In order to compare the NER performance of large language models in a zero-shot setting with that of a smaller model fine-tuned on domain-specific data, a training dataset was extracted from two separate sections of the Zibaldone (pp.1000-2001 and pp. 3001-4000) by sampling notes with a total length less than or equal to 350 tokens and which contained at least a reference annotated. The reason to filter out longer texts is to let the model focus on documents in which information is concentrated in small sequences, since the Zibaldone is often composed of short notes which are semantically independent. After this filtering strategy, we obtained a total of 688 notes which were processed following the same pipeline described above. Table \ref{tab3} reports the number of references in both the training and the evaluation dataset divided by class.  In total, the dataset contains 2,899 references to people, places and literary works in the Zibaldone annotated by domain experts.  The complete dataset is available on Zenodo~\cite{santini_zibaldoned_2024}.

In spite of the fact that this dataset has been sampled from a single author and that the NER annotations are limited to 3 coarse-grained classes, it still represents a valuable resource to test NER models on historical Italian for many reasons. First, the dataset contains ca. 850 references to literary works, which are one of the most challenging entity types to recognize in historical texts. Secondly, the data presents many domain-specific challenges from a linguistic point of view: one of them is that notes on the Zibaldone often contain quotes or expressions in Latin, Greek and French. Moreover, references to entities may be given by Leopardi with abbreviations, such as “Cic.” for “Cicero” or “Rep.” for “De Republica”. This is a frequent feature in scholarly writings and makes this dataset qualitatively similar to the AJMC corpus, a NER benchmark for 19th century classical commentaries in English, French and German~\cite{romanello_named_2024}.

\begin{table}
\caption{Sample of a csv containing notes from DigitalZibaldone.}
\label{tab1}
\centering
\begin{tabularx}{\textwidth}{|l|X|}
\hline
\textbf{doc\_id} & \textbf{text} \\
\hline
https://digitalzibaldone.net/node/p2721\_1 & Anche il Gelli confessava (ap. Perticari Degli Scritt. del Trecento l. 2. c. 13. p. 183.) che la lingua toscana non era stata applicata alle scienze. (24. Maggio 1823.). \\
\hline
\end{tabularx}
\end{table}

\begin{table}
\caption{Sample of a csv containing entity annotations.}
\label{tab2}
\centering
    \resizebox{\textwidth}{!}{
    \begin{tabular}{|l|l|l|l|l|l|}
    \hline
    \textbf{doc\_id}& \textbf{surface}& \textbf{start\_pos}& \textbf{end\_pos}& \textbf{identifier}& \textbf{type}\\
    \hline
    https://digitalzibaldone.net/node/p2721\_1 & Gelli & 9 & 14 & Q518160 & PER \\
    \hline
    https://digitalzibaldone.net/node/p2721\_1 & Perticari & 31 & 40 & Q3769747 & PER \\
    \hline
    https://digitalzibaldone.net/node/p2721\_1 & Degli Scritt. del Trecento & 41 & 67 & viaf34613848 & WORK \\
    \hline
    \end{tabular}
    }
\end{table}

\begin{table}
\caption{Number of annotations divided by class in the dataset.}
\label{tab3}
\centering
\begin{tabular}{|l|l|l|l|}
\hline
\textbf{Dataset} & \textbf{PER} & \textbf{LOC} & \textbf{WORK} \\
\hline
Training & 1,093 & 407 & 635 \\
\hline
Testing & 492 & 61 & 211 \\
\hline
\end{tabular}

\end{table}

\subsection{NER Algorithms}

\subsubsection{Instruction-tuned LLM}

The official LLaMa3.1 release provided by Meta was used in the experiments. In order to run the LLM on the setup configuration available, we used the 8B parameters instruction-tuned model available on Huggingface Transformers\footnote{\url{https://huggingface.co/meta-llama/Meta-Llama-3.1-8B-Instruct}}. Since this model is not per se a NER model, we exploited the \textit{in-context learning} capabilities of the conversational agent in order to generate answers which could be used to extract entities from a text. This was done by using two prompts:

\begin{itemize}
    \item \textbf{Generative prompt}: the goal is to write a new version of the text with people, locations, and literary works which are annotated within the text with the following pattern: \textit{left context <type>surface form</type> right context}
    \item \textbf{Extractive prompt}: given a passage from the Zibaldone, generates an ordered list of entities to be extracted from the text following the pattern: \textit{<type 1>surface form 1</type 1>, <type 2>surface form 2</type 2>}
\end{itemize}

The instructions for both prompts were fed in Italian to mantain coherence with the input text. For more details, notebooks which show the step by step procedure to generate the output with the above mentioned prompts are available on GitHub\footnote{\url{https://github.com/sntcristian/ZibaldoNER}}. 


After generating a response with the two prompts for each passage in our evaluation dataset, a post-processing algorithm was used to convert the output of the LLM into a CSV format which allows for automatic evaluation. First, a regular expression was used to identify from the answer all the annotations provided within angle brackets in order to get a list of pairs \textit{(type, surface form)}. Then, a filtering algorithm was used to remove all the pairs which contained additional entity types, such as \texttt{"DATE"} or \texttt{"ORG"}. Finally, the list of annotations was mapped to the start and end position of the tokens in the original passage and each annotation is converted to a tuple \textit{(Note ID, Surface Form, Start, End, Type)}.

\subsubsection{Domain-specific NER Model}

The GliNER library\footnote{\url{https://pypi.org/project/gliner/}} was used to compare LLaMa3.1 with a smaller domain-specific NER model fine-tuned on our dataset. GliNER~\cite{zaratiana_gliner_2024} are a family of NER models based on BERT~\cite{devlin_bert_2019} which were proposed as more efficient alternatives to large autoregressive models such as LLaMa~\cite{touvron2023llamaopenefficientfoundation}. This architecture aims to encode both spans (i.e., tokens) and entity representations in a shared embeddings space and the learning objective is to maximize the dot product between the span and class vectors. One of the advantages of these models is to be able to perform NER in a zero-shot setting by detecting entities of unseen classes.

In order to train a domain-specific model, a GliNER model trained for general-purpose NER in Italian with 90M parameters was used as base variant\footnote{\url{https://huggingface.co/DeepMount00/GLiNER_ITA_BASE}}. Thus, this model was trained in a cross-domain setting by fine-tuning it on the training portion of the Zibaldone. In order to carry the fine-tuning step, the dataset was pre-processed in order to convert entity types into the equivalent Italian words, e.g., "persona", "luogo" and "opera". After the pre-processing step, the dataset was split into training and validation using a $9/1$ ratio. The training was carried for four epochs with a learning rate of $5 \times 10^{-6}$ for the NER components, i.e. the feed forward neural network and the span representations, a learning rate of $1 \times 10^{-5}$ for the Transformer backbone, batch size $4$ and weight decay $0.01$. The low learning rate for the NER components was chosen in order to avoid catastrophic forgetting due to continual learning~\cite{chen2018continual}. The training was conducted on a Dell7920 machine equipped with an Nvidia RTX A6000 GPU.

Furthermore, to understand the advantages of training a domain-specific model with manually annotated data, we compared the fine-tuned GliNER model with its base variant in a zero-shot evaluation. Since GliNER models allow for generalized NER, which is the task of recognizing entity without a closed vocabulary of entity types, we tested the 90M parameter model for Italian in a zero-shot setting by tuning it in order to detect entities described by the class "person", "place" and "work".

\subsection{Evaluation Setup}
Named Entity Recognition (NER) performance was assessed using both exact and fuzzy matching criteria, which is a common practice in evaluating NER models~\cite{gonzalez-gallardo_injecting_2023, gonzalez-gallardo_yes_2023, gonzalez-gallardo_leveraging_2024}. Both approaches require that the predicted class matches the reference class, but they differ in how they handle span detection. The exact matching criterion considers an annotation correct only when the predicted tokens perfectly aligns with the gold standard, while the fuzzy matching criterion allows for partial overlaps between the predicted and gold standard tokens. Using both criteria offers additional insights: 
\begin{enumerate}
    \item it helps assess how well the algorithms return correct annotations even if the boundaries of the mention are not perfectly identified;
    \item examining the difference between exact and fuzzy matching in a per-class analysis allows to understand which entity types are more affected by boundary detection errors.
\end{enumerate}
For both criteria (exact and fuzzy), precision, recall, and F1-score are calculated, with results being micro and macro-averaged across all classes as well as computed separately for each class.

\section{Results}
\label{sec:results}

The results of the experiments are summarized in Table 4 and Table 5 respectively for the micro-averaged precision, recall and F1-scores and for all the metrics computed separately for each class and macro-averaged. The performance of four different algorithms - LLaMa3.1 (generative and extractive), GliNER (zero-shot and fine-tuned) - was evaluated using both exact and fuzzy matching. Both tables show that the GliNER fine-tuned model significantly outperforms the others across all metrics, achieving satisfactory results with a micro-averaged F1-score of 68.98\% (exact) and 75.64\% (fuzzy) on the overall dataset. This is particularly evident in the "person" class, where the model reaches a precision of 89,75\% (exact) and 92\% (fuzzy). 

\begin{table}[ht]
\label{tab4}
\centering
\begin{tabularx}{\textwidth}{l|XXX|XXX}
\toprule
\multicolumn{1}{c}{} & \multicolumn{3}{c|}{\textbf{Exact}} & \multicolumn{3}{c}{\textbf{Fuzzy}} \\
\midrule
          & \textbf{Precision} &  \textbf{Recall} &  \textbf{F1} &  \textbf{Precision} &  \textbf{Recall} &  \textbf{F1} \\
\midrule
LLaMa3.1-8B (generative) &  22,48 &            48,42 &        30,71 &               24,73 &            53,27 &        33,78 \\
LLaMa3.1-8B (extractive) &      \underline{37,06} &            29,06 &        32,58 &               \underline{44,07} &            34,55 &        38,74 \\
GliNER (zero-shot)  &    30,6 &            \underline{50,79} &        \underline{38,19} &                35,33 &            \underline{58,64} &        \underline{44,09} \\
GliNER (fine-tuned) &   \textbf{75,15} &            \textbf{63,74} &        \textbf{68,98} &               \textbf{82,4} &            \textbf{69,9 }&        \textbf{75,64} \\
\bottomrule
\end{tabularx}
\caption{Micro-averaged results of precision, recall and F1-score of the four algorithms computed on the evaluation dataset. For each evaluation metric, bold and underlined represent best and second best performance respectively.}
\end{table}

\begin{table}[ht]
\label{tab5}
\centering
\begin{tabularx}{\textwidth}{X|l|XXXX|XXXX}
\toprule
\multicolumn{2}{c}{} & \multicolumn{4}{c|}{\textbf{Exact}} & \multicolumn{4}{c}{\textbf{Fuzzy}} \\
\midrule
\multicolumn{2}{c|}{} & \textbf{PER} &  \textbf{LOC} &  \textbf{WORK} &  \textbf{Avg.} &  \textbf{PER} &  \textbf{LOC} &  \textbf{WORK} &  \textbf{Avg.} \\
\midrule
\multirow{4}{*}{\rotatebox{90}{Precision}}   & LLaMa3.1-8B (generative) & 28,97 & 9,00 & 12,69 & 16,89 & 29,85 & 9,00 & 18,07 & 18,97\\ 
                        & LLaMa3.1-8B (extractive) & \underline{56,87} & \underline{15,85} & 15,19 & \underline{29,30} & \underline{61,02} & \underline{17,07} & 28,92 & \underline{35,67}\\ 
                        & GliNER (zero-shot) & 45,18 & 12,96 & \underline{15,86} & 24,67 & 46,68 & 14,07 & \underline{29,94} & 30,23\\ 
                        & GliNER (fine-tuned) & \textbf{89,75} & \textbf{81,25} & \textbf{44,50} & \textbf{71,83} & \textbf{92,00} & \textbf{81,25} & \textbf{63,50} & \textbf{78,92}\\ \midrule
\multirow{4}{*}{\rotatebox{90}{Recall}}   & LLaMa3.1-8B (generative) & 59,76 & 16,39 & \underline{31,28} & 35,81 & 61,59 & 16,39 & 44,55 & 40,84\\ 
                        & LLaMa3.1-8B (extractive) & 36,18 & 21,31 & 14,69 & 24,06 & 38,82 & 22,95 & 27,96 & 29,91\\ 
                        & GliNER (zero-shot) & \underline{60,98} & \underline{57,38} & 25,11 & \underline{47,82} & \underline{63} & \underline{62,29} & \underline{47,39} & \underline{57,56}\\ 
                        & GliNER (fine-tuned) & \textbf{72,97} & \textbf{63,93} & \textbf{42,18} & \textbf{59,69} & \textbf{74,80} & \textbf{63,93} & \textbf{60,19} & \textbf{66,31}\\ \midrule
\multirow{4}{*}{\rotatebox{90}{F1}}   & LLaMa3.1-8B (generative) & 39,02 & 11,63 & 18,06 & 22,9 & 40,21 & 11,63 & 25,72 & 25,85\\ 
                        & LLaMa3.1-8B (extractive) & 44,22 &  18,18 & 14,94 & 25,78 & 47,45 & 19,58 & 28,43 & 31,82\\ 
                        & GliNER (zero-shot) & \underline{51,9} & \underline{21,15} & \underline{19,45} & \underline{30,83} & \underline{53,63} & \underline{22,96} & \underline{36,7} & \underline{37,76}\\ 
                        & GliNER (fine-tuned) & \textbf{80,49} & \textbf{71,56} & \textbf{43,3} & \textbf{65,11} & \textbf{82,51} & \textbf{71,56} &            \textbf{61,8} & \textbf{71,96}\\ \hline
\end{tabularx}
\caption{Precision, Recall and F1-scores computed per class and macro-averaged in both exact and fuzzy matching settings. For each class, bold and underlined represent best and second best performance respectively.}
\end{table}

On the other hand, the LLaMa3.1 model with a generative prompt shows the weakest performance with an overall F1-score of 30.71\% (exact) and 33.78\% (fuzzy). This indicates that the generative capabilities of this model are not well-suited for NER tasks in historical texts, particularly when it comes to exact span detection. However, it is to notice that there is a slight difference in precision and recall when comparing the generative and extractive approaches, with the second achieving better precision. Indeed, in a per-class analysis, it is remarkable how LLaMa3.1 with an extractive prompt achieves the second best performance in terms of precision for both "person" and "location", outperforming the base variant of GliNER.

In terms of entity classes, the "person" class generally exhibits higher scores across models, with GliNER fine-tuned showing the best results with an F1-score of 82,51\% and a precision of 92\% (both fuzzy). However, the "work" class continues to present challenges, especially for non-trained models like LLaMa and GliNER zero-shot, where F1 drops below 20\% in the exact matching setting. Additionally, the variance between exact and fuzzy matching is most pronounced in this class, indicating that models struggle to correctly identify boundaries for creative works, which often results in partial matches rather than exact ones.

\section{Discussion and Conclusion}
\label{sec:discussion_and_conclusion}

\subsection{Discussion}
The performance of the models shows a clear divide between those that are fine-tuned and those that rely solely on in-context learning or zero-shot capabilities. GliNER fine-tuned consistently outperforms the other models, demonstrating the value of domain-specific training when applied to complex literary texts. Its ability to identify entities with high precision and recall in both the "person" and "place" classes supports this observation. Particularly, for the "person" class, it reaches a precision of 92\% in the fuzzy matching criterion, indicating robustness even in challenging text spans such as abbreviations. Conversely, the LLaMa3.1 model with a generative prompt emerges as the least performing model, with an F1-score of 30.71\% (exact). This suggests that LLMs, even with their large parameter bases, may not be well-suited for tasks requiring precise entity recognition without fine-tuning, particularly in historical or literary contexts.

Among the entity classes, unexpectedly, the "place" class proves to be the most difficult to predict across LLMs. Instead models which are trained for general-domain NER, such as GliNER applied in a zero-shot setting, are achieving slightly better performance. This may be due to the fact that toponyms in Leopardi's notes may present lexical variations such as no-capitalized words, e.g. "italia", or abbreviations, e.g. "Venez." for Venice. Overall, the most challenging class, even for the best performing model, is the "work" class. This is a well-known problem in NER, due to the complex nature of references to creative works, which may often present abbreviations and lexical variations \cite{jain2019mona}. Specifically, a common error across all three models is the inability to distinguish bibliographic references expressed through the reference to the author of the work itself, which is a special ambiguity which is exemplified by Leopardi's frequent use of "Il Forcellini" to refer to the "Lexicon Totius Latinitatis" (1771). Such references require contextual information often absent in language models, leading them to classify these portions of text as references to people rather than literary works. Moreover, the "work" class exhibits the highest variance between exact and fuzzy matching, particularly in models like LLaMa3.1. This suggests that the boundaries for creative works are often misidentified, possibly due to the complexity of literary references and the inherent ambiguity of titles within the text. Fuzzy matching alleviates some of these issues, but precise identification remains a challenge.

\subsection{Conclusion}
In summary, this research proposed a comparative analysis of four NER algorithms based on language models for historical Italian literary texts. For this purpose, a new dataset was proposed that can be used by the entire research community to test the performance of various automated systems not only for NER but also for linking these references to Wikidata, i.e., Entity Linking (EL). This dataset, collected through the scraping of the online resource DigitalZibaldone, aims to become a new benchmark for measuring the effectiveness of automated systems supporting the creation of digital editions. Additionally, this study presents the application of four different NER approaches on this dataset and a quantitative analysis of the NER results obtained through standard metrics in information retrieval. The main takeaway from this research is that dealing with historical literary texts like Leopardi's Zibaldone presents unique challenges for NER tasks. One key difficulty is the recognition of domain-specific entities, particularly references to literary works, which are often referred to indirectly or with incomplete context. Models must also handle non-standard language use and orthographic variations that are typical of older texts, such as abbreviations and latin names.

In conclusion, what emerges from this research is that automatic entity recognition systems can be applied, albeit with several challenges, to humanistic texts. Fine-tuned models trained on domain data can be used to annotate literary texts, but human supervision is still necessary to verify the correctness and completeness of the annotations obtained. This suggests that NLP systems can be useful allies in digital editing projects, provided they are integrated into human-in-the-loop systems. On the other hand, large language models still seem far from being successfully applied to the analysis of humanistic texts, probably due to the predominance of web-derived texts in the training corpora of these models.
Future extensions of this work will focus on two main directions. The first will be to test models not only for entity recognition but also for EL, i.e., the disambiguation of a reference by linking it to its respective Wikidata element. This will aim to assess the feasibility of an architecture that automatically annotates references present in a literary corpus and semantically enriches them by linking them to an external \textit{ad-hoc} knowledge base. The second direction, closely related to the first, is to propose a methodology for correcting and improving the performance of predictive systems for NER and EL by integrating them with semantic and logical conditions as well as statistical approaches. Since Knowledge Bases such as Wikidata~\cite{vrandevcic2014wikidata} or Yago~\cite{suchanek2024yago} contain multiple attributes and properties describing specific resources such as people, places and books, this information can be used to justify or contradict the predictions of an automatic entity extraction system by establishing temporal, spatial, or logical conditions to filter the elements of a knowledge base, as already proposed in other studies~\cite{gonzalez-gallardo_injecting_2023}.

\bibliography{ner_bibliography}

\end{document}